\documentclass[lettersize,journal]{IEEEtran}
\usepackage{amsmath,amsfonts}
\usepackage{algorithmic}
\usepackage{algorithm}
\usepackage{array}
\usepackage{textcomp}
\usepackage{stfloats}
\usepackage{url}
\usepackage{verbatim}
\usepackage{graphicx}
\usepackage{subfigure}
\usepackage{caption, subcaption}
\usepackage{cite}
\usepackage{float}
\begin{document}
\title{Seg-CycleGAN : SAR-to-optical image translation guided by a downstream task}
\author{Hannuo Zhang, Huihui Li$^{\ast}$, Jiarui Lin, Yujie Zhang, Jianghua Fan, and Hang Liu
\thanks{*Corresponding author}
\thanks{Hannuo Zhang is with the the Gina Cody School of Engineering and Computer Science, Concordia University, Montreal H3G1M8, Canada.}
\thanks{Huihui Li, Jiarui Lin, Yujie Zhang, Jianghua Fan are with the School of Automation, Northwestern Polytechnical University, Xi'an 710072, China.}
\thanks{Hang Liu is with the School of Cybersecurity, Northwestern Polytechnical University, Xi'an 710072, China.}
}
\maketitle 
\begin{abstract}
Optical remote sensing and Synthetic Aperture Radar(SAR) remote sensing are crucial for earth observation, offering complementary capabilities. While optical sensors provide high-quality images, they are limited by weather and lighting conditions. In contrast, SAR sensors can operate effectively under adverse conditions. This letter proposes a GAN-based SAR-to-optical image translation method named Seg-CycleGAN, designed to enhance the accuracy of ship target translation by leveraging semantic information from a pre-trained semantic segmentation model. Our method utilizes the downstream task of ship target semantic segmentation to guide the training of image translation network, improving the quality of output Optical-styled images. The potential of foundation-model-annotated datasets in SAR-to-optical translation tasks is revealed. This work suggests broader research and applications for downstream-task-guided frameworks. The code will be available at \url{https://github.com/NPULHH/}.
\end{abstract}
\section{Introduction}
\IEEEPARstart{O}{ptical} remote sensing and Synthetic Aperture Radar(SAR) remote sensing are important means to capture images for earth observation. Although images acquired by optical remote sensing feature high quality, this method becomes ineffective under conditions like clouds, fog and nighttime. However, Synthetic Aperture Radar, as an active sensor, can function normally under nighttime and cloudy conditions. As remote sensing technology is a crucial measure for ensuring the safety of ships, effective monitoring and supporting rescue operations, transforming ship targets from SAR images to optical-styled images using Generative Adversarial Networks (GANs)\cite{GANs} has significant research value but also faces certain challenges. According to dependence on paired datasets, current GAN-based methods for SAR-to-optical image translation can be divided into paired and unpaired methods\cite{comparative_analysis}. Using paired datasets and corresponding paired image-to-image translation methods\cite{Pix2Pix} can impose more strict constraints on generator modules, which makes it more likely to achieve excellent translation results. However, for moving targets like ships, obtaining paired datasets is extremely difficult, leaving unpaired translation methods the only choice. Among unpaired methods, U-GAT-IT\cite{U_GAT_IT} introduces attention modules and an adaptive normalization function to achieve image translation requiring large shape changes. NICE-GAN\cite{NICE_GAN} achieves a compact model structure and more effective training by reusing the first few layers of its discriminator module as an image encoder. CUT\cite{CUT} applies contrastive learning to gain great performance on data that does not meet bijective relationship required by cycle-consistency loss\cite{CycleGAN}. However, in the field of SAR to optical translation, among unpaired methods, CycleGAN\cite{CycleGAN} has relatively better performance\cite{comparative_analysis}. As for target-level SAR-to-optical image translation, \cite{Aircraft_Translation} studied SAR-to-optical image translation for aircrafts, but they used CAD models to generate rendered optical-styled data paired with SAR images for training. Due to the existence of Sim2Real gap,  application value of this algorithm is limited. Besides, current translation methods are carried out independently of downstream tasks such as semantic segmentation, detection, or recognition of ships. Thus the abscence of constraints from downstream tasks in training makes it difficult for image translation models to learn to capture semantic information that distinguishes ship targets from background. This leads to poor performance in image translation tasks targeting specific objects like ships. Additionally, there are no suitable evaluation metrics to assess image translation performance from the perspective of downstream tasks.

Our contributions to address the above issues are as follows: A GAN-based SAR-to-optical image translation method driven by a downstream task is proposed to perform image translation for ship targets. Our method uses a pre-trained semantic segmentation module as part of the objective function. This module can guide the training of image translation with semantic information gained from the downstream task on optical data. By utilizing Segment Anything Model\cite{SAM} to annotate optical data\cite{SAMRS}, we obtained accurate ship segmentation labels to train a ship segmentation module, which makes it possible to guide the training of the SAR-to-optical translation model with semantic information related to ship targets. Experiments conducted on DIOR\cite{DIOR} and HRSID\cite{HRSID} datasets proves our method to be effective. Experiments conducted on WHU-OPT-SAR dataset\cite{WHU_OPT_SAR} further demonstrate that our method is also effective in scenes of different scales.

In the remaining parts of this letter, we first introduce our proposed Seg-CycleGAN method in section II. Results of experiments are presented in Section III. And finally, the conclusion is drawn in section IV.
\begin{figure*}[htbp]
\center{\includegraphics[width=15cm]  {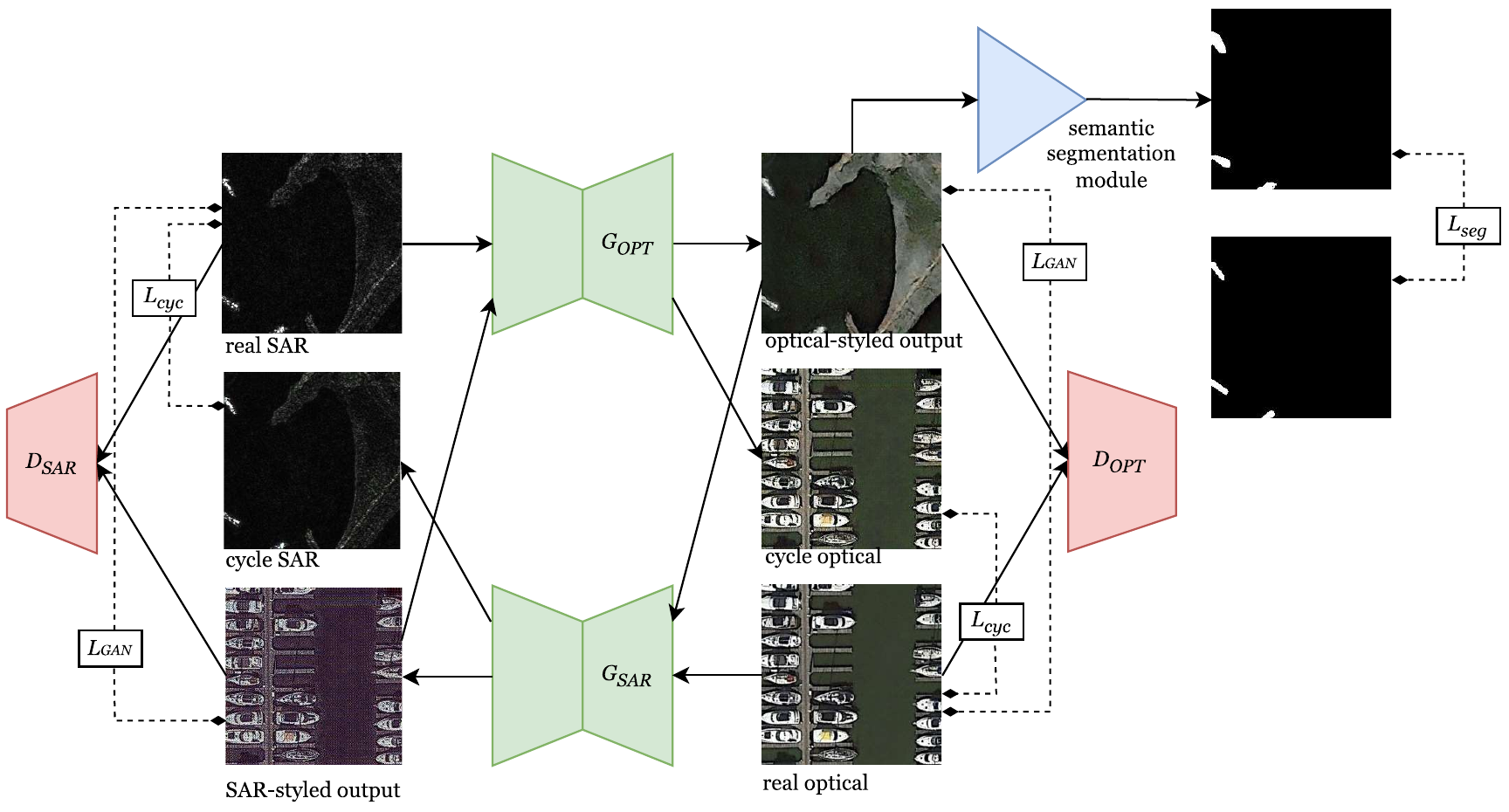}} 
	\caption{Structure of Seg-CycleGAN for SAR-to-optical image translation.}
	\label{fig_1}
\end{figure*}

\section{Method}
\subsection{Labeling Datasets with SAM}
It is challenging to find a semantic segmentation dataset for ship targets, but ship target detection datasets do exist. SAM\cite{SAM}, a Foundation Model in the field of semantic segmentation, has excellent zero-shot capability. According to the processing method mentioned in SAMRS\cite{SAMRS}, the HBB annotations from the ship detection dataset are used as prompts for SAM, and output of SAM is used as segmentation labels for ship targets. Due to significant differences in color and texture between ships and their background(such as water surface or shores), the segmentation annotations are relatively accurate. This method is applied to the HRSC2016-MS\cite{HRSC2016_MS} and DIOR datasets to obtain semantic segmentation datasets of ship targets. After appropriate data augmentation, the ship segmentation dataset is used to train a ship-specific binary semantic segmentation module. Images in acquired dataset are also used in training and testing of SAR-to-optical image translation models.

\subsection{Network Architectures}
Since there is no paired SAR-optical ship target dataset, and in the field of SAR-to-optical image translation, performance of CycleGAN is relatively better than other unpaired methods\cite{comparative_analysis}, CycleGAN architecture is adopted as the backbone network. Thus the backbone network contains 2 generator modules and 2 discriminator modules as Fig. 1 shows. The generators are used to perform image style transferring, taking real SAR images as input and outputting optical-styled images, or taking real optical images as input and outputting SAR-styled images, while the purpose of discriminators is to distinguish whether received data is real or generated. The generators and discriminators are trained alternately. In their competition-like training process, the ability of discriminators to distinguish real and generated data gradually improves, and the data distribution of output of generators becomes increasingly close to the real distribution. A semantic segmentation module is used to perform semantic segmentation, i.e. , assigning each pixel a label of either ship or background, on the optical-styled data generated by the SAR-to-optical generator. In our experiments, both generators of the Seg-CycleGAN method consist of 3 convolutional down-sampling layers, 9 residual connection blocks, and 2 deconvolutional up-sampling layers connected in sequence, followed by convolutional layers and a tanh activation function to obtain the final output. Each discriminator adopts the PatchGAN architecture\cite{Pix2Pix}, consisting of 5 convolutional layers with LeakyReLU activation function. SegNet structure\cite{SegNet} is adopted by the semantic segmentation module, which consists of an encoder, a decoder, and a pixel-level classifier connected in series. The encoder contains 13 convolutional layers, and the decoder consists of 13 nonlinear up-sampling layers.

\subsection{Loss Functions}
Adversarial loss\cite{GANs} ${{\cal L}_{GAN}}$ of classical generative adversarial networks and cycle-consistency loss ${{\cal L}_{cyc}}$ are adopted. ${D_{OPT}}$ is trained to distinguish the authenticity of optical-styled data, while ${D_{SAR}}$ is used to distinguish authenticity of SAR-styled data. $rs$ represents real SAR images, and $ro$ represents real optical data. ${G_{OPT}}$ is used to generate optical-styled data, and ${G_{SAR}}$ is trained to output SAR-styled images. The definitions of ${{\cal L}_{GAN}}$ and ${{\cal L}_{cyc}}$ are as follows:
\begin{equation}
\begin{split}
\label{L-GAN-OPT}
{{\cal L}_{GAN}}({G_{OPT}},{D_{OPT}}) = \\
{E_{ro}}[\log {D_{OPT}}(ro)] + {E_{rs}}[\log (1 - {D_{OPT}}({G_{OPT}}(rs)))]
\end{split}
\end{equation}
\begin{equation}
\begin{split}
\label{L-GAN-SAR}
{{\cal L}_{GAN}}({G_{SAR}},{D_{SAR}}) = \\
{E_{rs}}[\log {D_{SAR}}(rs)] + {E_{ro}}[\log (1 - {D_{SAR}}({G_{SAR}}(ro)))]
\end{split}
\end{equation}
\begin{equation}
\begin{split}
\label{L-CYC}
{{\cal L}_{cyc}}({G_{OPT}},{G_{SAR}}) = \\
{E_{rs}}[{\left\| {{G_{SAR}}({G_{OPT}}(rs))} \right\|_1}] + {E_{ro}}[{\left\| {{G_{OPT}}({G_{SAR}}(ro))} \right\|_1}]
\end{split}
\end{equation}
During the alternating training proccess, ${D_{OPT}}$ and ${D_{SAR}}$ are optimized to maximize ${{\cal L}_{GAN}}$, while ${G_{OPT}}$ and ${G_{SAR}}$ are optimized to minimize ${{\cal L}_{GAN}}$.  ${{\cal L}_{cyc}}$ requires that the generated data inputted to another generator produces an output that highly resembles the original input. While ${{\cal L}_{GAN}}$ facilitates bidirectional style transferring, ${{\cal L}_{cyc}}$ ensures that the image translation process retains information contained in the input image. Under the influence of ${{\cal L}_{cyc}}$, the SAR-to-optical generator used in experiments effectively preserves different regions in input SAR images, including land, sea and ships. One major advantage of GANs is using generators and discriminators to replace explicitly defined complex loss functions. Based on this idea, the pre-trained semantic segmentation module can be viewed as a loss function highly adapted to the data distribution of the target optical domain, e.g., segmentation loss ${{\cal L}_{seg}}$. This module is trained with a SAM-annotated ship target semantic segmentation dataset before the training of other modules of Seg-CycleGAN. When used to guide the training of ${G_{OPT}}$, parameters of the semantic segmentation module are fixed. During training, cross-entropy loss of generated optical-styled data and corresponding SAR image segmentation labels is calculated, and the obtained gradients during back propagation are used to update the weights of the SAR-to-optical generator. The pre-trained semantic segmentation module contains information for recognizing whether a pixel belongs to a ship or other scenes in optical images, including shape, color, and edge. Hence using it as a loss function encourages the generator working with unpaired data to convert ship regions in SAR images into ship regions in the optical-styled output, with other regions correspondingly converted to optical-styled sea or land, suppressing appearance of unreal targets and textures. The segmentation loss is defined as follows:
\begin{equation}
\begin{split}
\label{L-SEG}
{{\cal L}_{seg}}({G_{OPT}}) = Cross\_Entropy(gt\_seg,{S_\theta }({G_{OPT}}(rs)))
\end{split}
\end{equation}
where $gt\_seg$ represents the semantic segmentation labels for ships and ${S_\theta }$ represents the pre-trained semantic segmentation module parametrized by ${\theta}$. The definition of overall objective function ${\cal L}({G_{OPT}},{G_{SAR}},{D_{OPT}},{D_{SAR}})$ for the 2 generator modules in Seg-CycleGAN is as follows, composed of ${{\cal L}_{GAN}}$, ${{\cal L}_{cyc}}$, and ${{\cal L}_{seg}}$:
\begin{equation}
\begin{split}
\label{L-FULL}
{\cal L}({G_{OPT}},{G_{SAR}},{D_{OPT}},{D_{SAR}}) = \\
{{\cal L}_{GAN}}({G_{OPT}},{D_{OPT}}) + {{\cal L}_{GAN}}({G_{SAR}},{D_{SAR}}) + \\
\alpha {{\cal L}_{cyc}}({G_{OPT}},{G_{SAR}}) + \\
\beta {{\cal L}_{seg}}({G_{OPT}})
\end{split}
\end{equation}
where $\alpha$ and $\beta$ are hyperparameters used to balance different terms. In the experiments, $\alpha$ is set to $10.0$, and $\beta$ is set to $0.3$.

\subsection{Training Details}
\begin{figure*} 
    \subfigure[SAR]{
    \begin{minipage}[t]{0.16\linewidth} 
        \centering
        \vspace{3pt}
        \includegraphics[width=\textwidth]{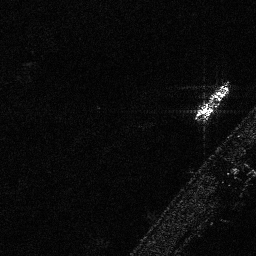}
        \vspace{3pt}
        \includegraphics[width=\textwidth]{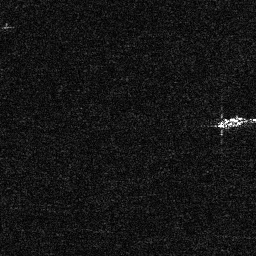}
        \vspace{3pt}
    \end{minipage}
    }%
    \subfigure[CUT]{
    \begin{minipage}[t]{0.16\linewidth} 
        \centering
        \vspace{3pt}
        \includegraphics[width=\textwidth]{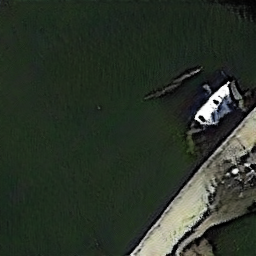}
        \vspace{3pt}
        \includegraphics[width=\textwidth]{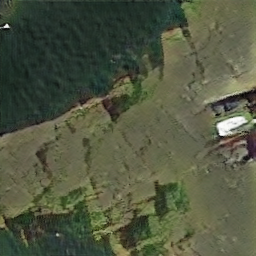}
        \vspace{3pt}
    \end{minipage}
    }%
    \subfigure[CycleGAN]{
    \begin{minipage}[t]{0.16\linewidth} 
        \centering
        \vspace{3pt}
        \includegraphics[width=\textwidth]{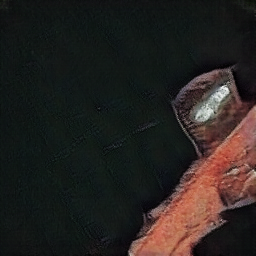}
        \vspace{3pt}
        \includegraphics[width=\textwidth]{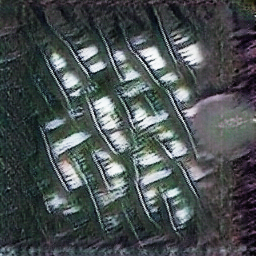}
        \vspace{3pt}
    \end{minipage}
    }%
    \subfigure[NICE-GAN]{
    \begin{minipage}[t]{0.16\linewidth} 
        \centering
        \vspace{3pt}
        \includegraphics[width=\textwidth]{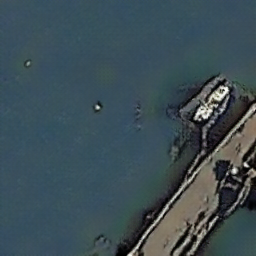}
        \vspace{3pt}
        \includegraphics[width=\textwidth]{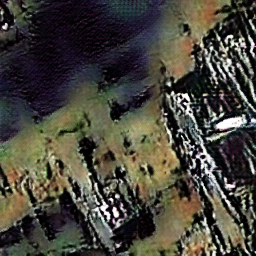}
        \vspace{3pt}
    \end{minipage}
    }%
    \subfigure[U-GAT-IT]{
    \begin{minipage}[t]{0.16\linewidth} 
        \centering
        \vspace{3pt}
        \includegraphics[width=\textwidth]{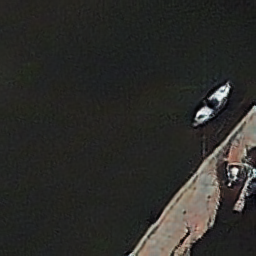}
        \vspace{3pt}
        \includegraphics[width=\textwidth]{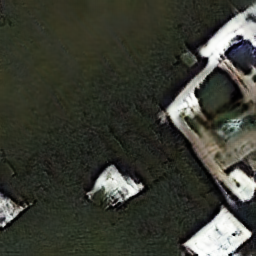}
        \vspace{3pt}
    \end{minipage}
    }%
    \subfigure[Seg-CycleGAN]{
    \begin{minipage}[t]{0.16\linewidth} 
        \centering
        \vspace{3pt}
        \includegraphics[width=\textwidth]{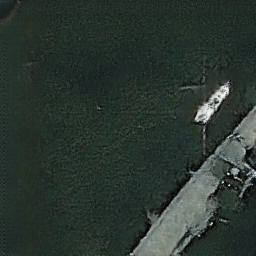}
        \vspace{3pt}
        \includegraphics[width=\textwidth]{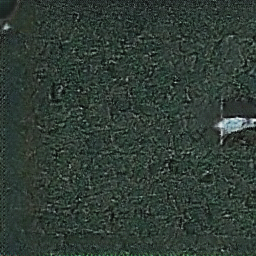}
        \vspace{3pt}
    \end{minipage}
    }%
    \centering
    \caption{Examples of ship target translation for samples in HRSID}
    \label{fig_2}
\end{figure*}
Before all experiments, we preprocessed datasets, excluding images without ships in the DIOR dataset and cropping samples from the HRSID and DIOR datasets. Original size of all samples is 800 x 800 pixels, which are cropped to 256 x 256 pixels with a step size of 205. After cropping, samples in HRSID with less than 95 ship pixels were removed. We augmented the optical dataset by rotating the DIOR samples at 10, 30, 50, 70, ..., 170 degrees for a total of 9 different angles. The SAM model version used in the experiments is ViT-H. Class weight ratio for cross-entropy loss during training of the semantic segmentation module is set to 0.143(background) : 0.857(ship)\cite{optical_ship_segmentation}. During calculation of segmentation loss, ship and background classes are weighted equally. We selected 1000 samples from WHU-OPT-SAR dataset and corrected their semantic segmentation annotation. For experiments on refined WHU-OPT-SAR dataset, the number of epochs is set to 200, for experiments on DIOR and HRSID, the number of epochs is set to 100. In all experiments, batch size is set to 1, and the Adam optimizer is used. All experiments were conducted on two GeForce GTX 1080 Ti GPUs.

\section{Experiment Results}
In this section, we evaluate experiments conducted on HRSID, DIOR, and refined WHU-OPT-SAR dataset. HRSID and DIOR together form an unpaired SAR-optical image dataset, which we refer to as HRSID-DIOR. On HRSID-DIOR, effectiveness of our proposed Seg-CycleGAN method in translating remote sensing data containing ships is proved. Performance of Seg-CycleGAN on land use classification datasets is also evaluated with WHU-OPT-SAR dataset.
\subsection{Datasets}
Experiments are conducted on subsets of the HRSID, DIOR, and WHU-OPT-SAR datasets. The HRSID dataset contains 5604 SAR images with ship targets, including 16951 ship instances, covering resolutions of 0.5m, 1m, and 3m, and is divided into nearshore and offshore parts. To obtain a higher proportion of data with complex scenes, we selected 1031 images from the nearshore part, 662 samples of which used as the training set for experiments on various unpaird image-to-image translation models. The DIOR dataset contains 23463 remote sensing images with 190288 target instances annotated in HBB format, covering 20 common object categories with spatial resolutions ranging from 0.5m to 30m.We selected data from 1302 training samples containing ship targets for training and 1900 testing samples containing ship targets for testing. The WHU-OPT-SAR dataset is a land use classification dataset containing 100 pairs of 5556x3704 pixel SAR-Optical pixel-registered images with a resolution of 5m, and it has segmentation labels for 8 types of land cover. We cropped these to 256x256 pixel samples and extracted 900 pairs for training and 198 pairs for testing in experiments on Pix2Pix and various unpaired models. For fairness, original paired SAR and optical images of WHU-OPT-SAR are randomly shuffled during the training of unpaired models.
\subsection{Qualitative Evaluation}
As shown in Fig. 2, the proposed Seg-CycleGAN performs well on both offshore data with simple images and nearshore data with more high-frequency information, accurately translating images containing ships at the pixel level. Even without segmentation loss constraints for translation of sea and land, they are accurately translated with reasonable colors and textures. Other unpaired methods, due to the lack of  pixel-level constraints, fail to achieve pixel-wise translation even when training loss is sufficiently lowered. For these methods, influenced by high-frequency information samples that contain shorelines or buildings in the training set, inputting offshore data generates complex, meaningless textures similar to docks and small boats. However, these methods learn the mapping between nearshore SAR images and nearshore optical images, resulting in relatively accurate output when inputting nearshore data. 
Experiments on refined WHU-OPT-SAR are also conducted, results of which are displayed in Fig. 3 of appendix. A binary semantic segmentation module is trained using farmland segmentation labels to guide the optimization of a SAR-to-optical translation model adopting CycleGAN structure. Application of segmentation loss successfully avoids mistaking farmland regions in SAR images for forests, which can happen to unpaired methods. The visual effects of translation results is significantly improved compared to other unpaired methods for the farmland category.
\subsection{Quantitative Evaluation}
\begin{table}[ht]
\centering
\begin{tabular}{|l|c|c|c|}
\hline
Method & \multicolumn{3}{c|}{segmentation-related metrics} \\
\hline
 & mPA & mIoU & FwIoU \\
\hline
Seg-CycleGAN & 0.824 & 0.678 & 0.933 \\
CycleGAN & 0.802 & 0.661 & 0.924 \\
NICE-GAN & 0.681 & 0.573 & 0.930 \\
U-GAT-IT & 0.724 & 0.599 & 0.931 \\
CUT & 0.754 & 0.629 & 0.922 \\
Ground Truth SAR & 0.734 & 0.652 & 0.898 \\
\hline
\end{tabular}
\caption{Quantitative Result for HRSID-DIOR - segmentation-related metrics}
\label{tab:segmentation-metrics}
\end{table}
\begin{table}[ht]
\centering
\begin{tabular}{|l|c|}
\hline
Method & FID \\
\hline
Seg-CycleGAN & 146.395 \\
CycleGAN & 159.319 \\
NICE-GAN & 153.703 \\
U-GAT-IT & 124.213 \\
CUT & 109.564 \\
\hline
\end{tabular}
\caption{Quantitative Result for HRSID-DIOR - similarity-related metrics}
\label{tab:similarity-metrics}
\end{table}
\begin{table}[ht]
\centering
\begin{tabular}{|l|c|c|c|c|}
\hline
Method & \multicolumn{4}{c|}{similarity-related metrics} \\
\hline
 & PSNR & SSIM & FID & cosine similarity \\
\hline
Seg-CycleGAN & 19.259 & 0.346 & 132.68 & 0.974 \\
CycleGAN & 18.748 & 0.339 & 132.926 & 0.974 \\
NICE-GAN & 19.219 & 0.304 & 167.523 & 0.971 \\
U-GAT-IT & 18.400 & 0.313 & 165.206 & 0.968 \\
CUT & 17.500 & 0.281 & 156.847 & 0.970 \\
Pix2Pix & 22.212 & 0.373 & 155.926 & 0.978 \\
\hline
\end{tabular}
\caption{Quantitative Result for WHU-OPT-SAR - similarity-related metrics}
\label{tab:whu-similarity-metrics}
\end{table}
To evaluate different algorithms from the perspective of image similarity, we use 4 metrics: PSNR, SSIM, cosine similarity, and FID. To measure the accuracy of translated ship targets, we applied 3 evaluation metrics originally used for semantic segmentation: mAP, mIoU, and FwIoU. These metrics are measured using the following steps:
\begin{enumerate}
\item{Complete the training of different SAR-to-optical image translation models.}
\item{Use the generators of trained SAR-to-optical image translation models to convert subsets of the HRSID into corresponding optical-styled images.
}
\item{Train semantic segmentation models adopting SegNet structure using the generated optical-styled training sets obtained from the second step and corrsponding segmentation labels.}
\item{Use trained SegNet models from the previous step for inference on generated optical-styled testing sets and measure the segmentation-related metrics.}
\end{enumerate}
Table I assesses 5 unpaired methods on 3 segmentation-related metrics in the experiments on HRSID-DIOR dataset, which also contains values of segmentation-related metrics for training SegNet directly on corresponding SAR images. From Table I, it is observed that Seg-CycleGAN achieves the best results on all metrics due to guidance of semantic segmentation module. Table II assesses performance of different algorithms with FID on HRSID-DIOR dataset. It can be observed that the proposed Seg-CycleGAN performs better than CycleGAN and NICE-GAN, while CUT and U-GAT-IT achieve lower FID values. Although the translation results of CUT and U-GAT-IT are closer in color to the real optical data in the DIOR dataset, their translation results are extremely inaccurate, with Seg-CycleGAN achieving the best overall visual effect. Table III evaluates performance of Pix2Pix and 5 unpaired methods on the WHU-OPT-SAR dataset. It is observed that the Pix2Pix method using paired data achieves the best results on most metrics. PSNR and SSIM metrics of the Seg-CycleGAN method are second only to Pix2Pix. For cosine similarity, performance of Seg-CycleGAN is close to CycleGAN and Pix2Pix. The above experiments indicate that under the constraints of unpaired data, Seg-CycleGAN can generate accurate and visually superior translation results of SAR images.
\section{Conclusion}
This letter proposes a SAR-to-optical image translation method driven by a downstream task, named as Seg-CycleGAN. Its advantage lies in using a pre-trained semantic segmentation module to enable the generator of GAN-based algorithms to perform accurate translation of SAR images containing ship targets. Additionally, this letter demonstrates the high application value of datasets annotated by SAM in tasks of semantic segmentation and image-to-image translation. Experiments on a land use classification dataset and ship target datasets prove that Seg-CycleGAN can be applied to SAR-to-optical image translation task of different resolutions and scenarios. We hope this letter can promote the research and application of downstream-task-guided translation frameworks.
\bibliographystyle{IEEEtran}
\bibliography{references/all_references}
\newpage
\onecolumn
\appendix

\section{Appendix}
\subsection{Visualization of experiments on WHU-OPT-SAR}
Fig. 3 shows translation results of different algorithms on WHU-OPT-SAR.
\begin{figure}[h]
    \centering
    \subfigure[SAR]{
    \begin{minipage}[t]{0.24\linewidth} %
        \centering
        \vspace{3pt}
        \includegraphics[width=\textwidth]{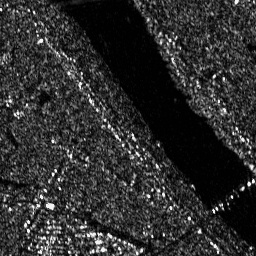}
        \vspace{3pt}
        \includegraphics[width=\textwidth]{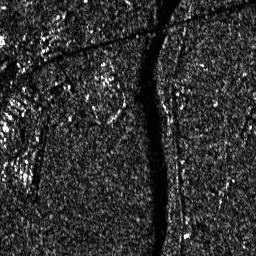}
        \vspace{3pt}
    \end{minipage}
    }%
    \subfigure[Pix2Pix]{
    \begin{minipage}[t]{0.24\linewidth} %
        \centering
        \vspace{3pt}
        \includegraphics[width=\textwidth]{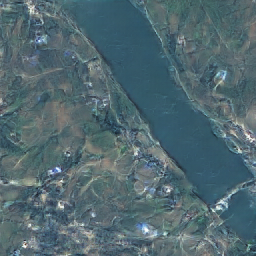}
        \vspace{3pt}
        \includegraphics[width=\textwidth]{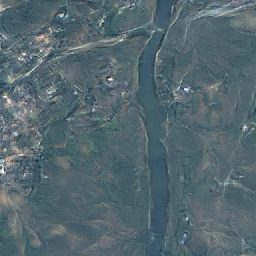}
        \vspace{3pt}
    \end{minipage}
    }%
    \subfigure[CycleGAN]{
    \begin{minipage}[t]{0.24\linewidth} %
        \centering
        \vspace{3pt}
        \includegraphics[width=\textwidth]{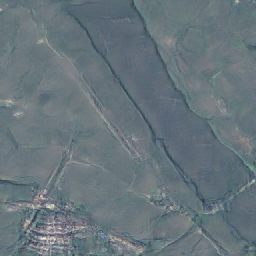}
        \vspace{3pt}
        \includegraphics[width=\textwidth]{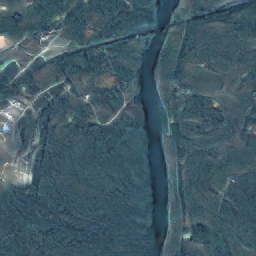}
        \vspace{3pt}
    \end{minipage}
    }%
    \subfigure[CUT]{
    \begin{minipage}[t]{0.24\linewidth} %
        \centering
        \vspace{3pt}
        \includegraphics[width=\textwidth]{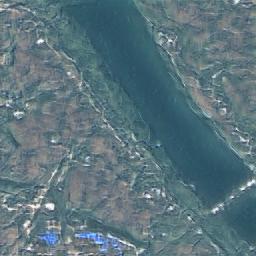}
        \vspace{3pt}
        \includegraphics[width=\textwidth]{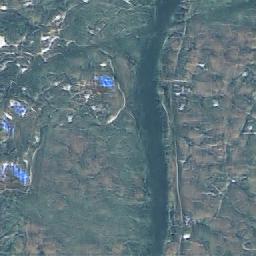}
        \vspace{3pt}
    \end{minipage}
    }%

    \subfigure[NICE-GAN]{
    \begin{minipage}[t]{0.24\linewidth} %
        \centering
        \vspace{3pt}
        \includegraphics[width=\textwidth]{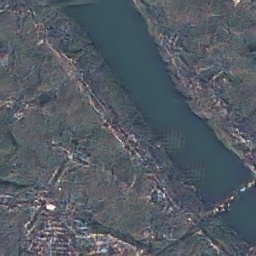}
        \vspace{3pt}
        \includegraphics[width=\textwidth]{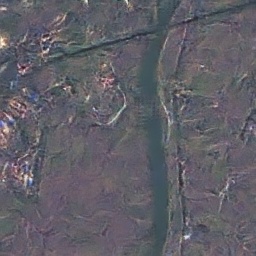}
        \vspace{3pt}
    \end{minipage}
    }%
    \subfigure[U-GAT-IT]{
    \begin{minipage}[t]{0.24\linewidth} %
        \centering
        \vspace{3pt}
        \includegraphics[width=\textwidth]{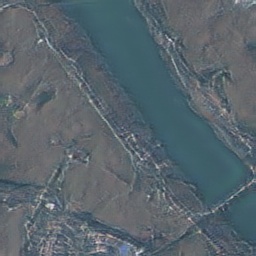}
        \vspace{3pt}
        \includegraphics[width=\textwidth]{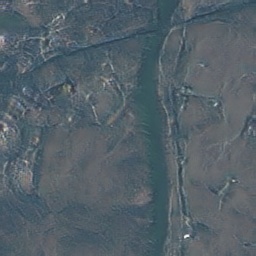}
        \vspace{3pt}
    \end{minipage}
    }%
    \subfigure[Seg-CycleGAN]{
    \begin{minipage}[t]{0.24\linewidth} %
        \centering
        \vspace{3pt}
        \includegraphics[width=\textwidth]{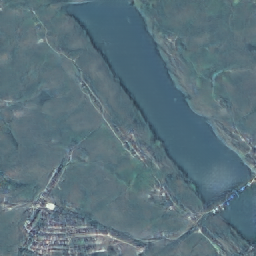}
        \vspace{3pt}
        \includegraphics[width=\textwidth]{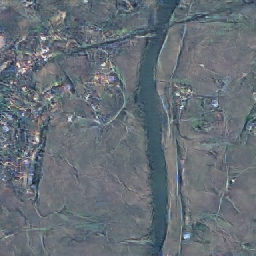}
        \vspace{3pt}
    \end{minipage}
    }%
    \subfigure[Optical]{
    \begin{minipage}[t]{0.24\linewidth} %
        \centering
        \vspace{3pt}
        \includegraphics[width=\textwidth]{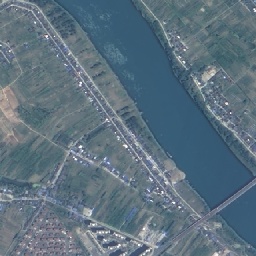}
        \vspace{3pt}
        \includegraphics[width=\textwidth]{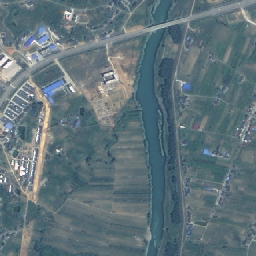}
        \vspace{3pt}
    \end{minipage}
    }%
    \centering
    \caption{Examples of image translation for samples in WHU-OPT-SAR}
    \label{fig_3}
\end{figure}
\newpage
\subsection{Visualization of SAM-annotated datasets}
We present some samples of SAM-annotated datasets, as shown in Fig. 4-5.
\begin{figure}[h] 
    \centering
    \subfigure[images]{
    \begin{minipage}{0.35\linewidth} 
        \centering
        \includegraphics[width=\textwidth]{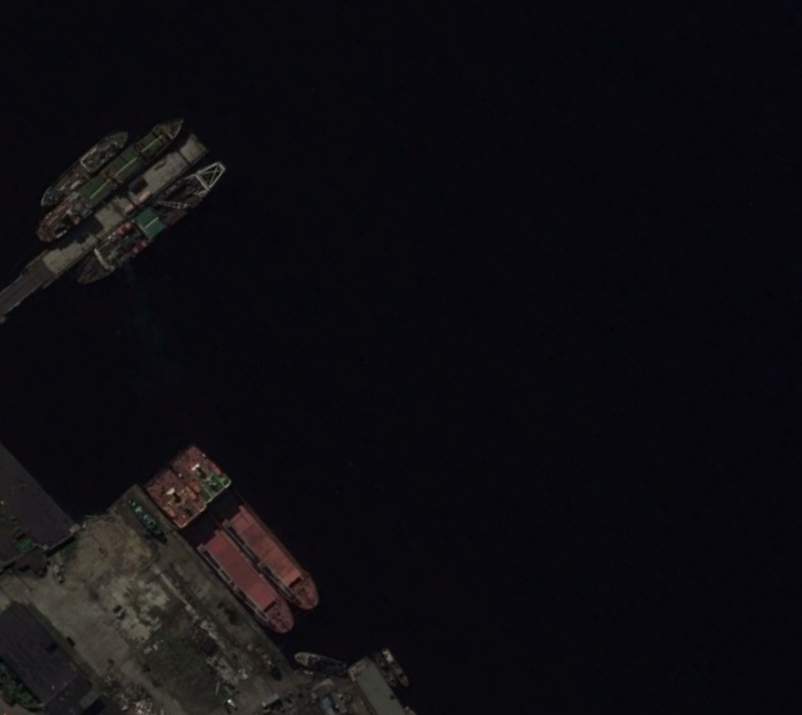}
        \includegraphics[width=\textwidth]{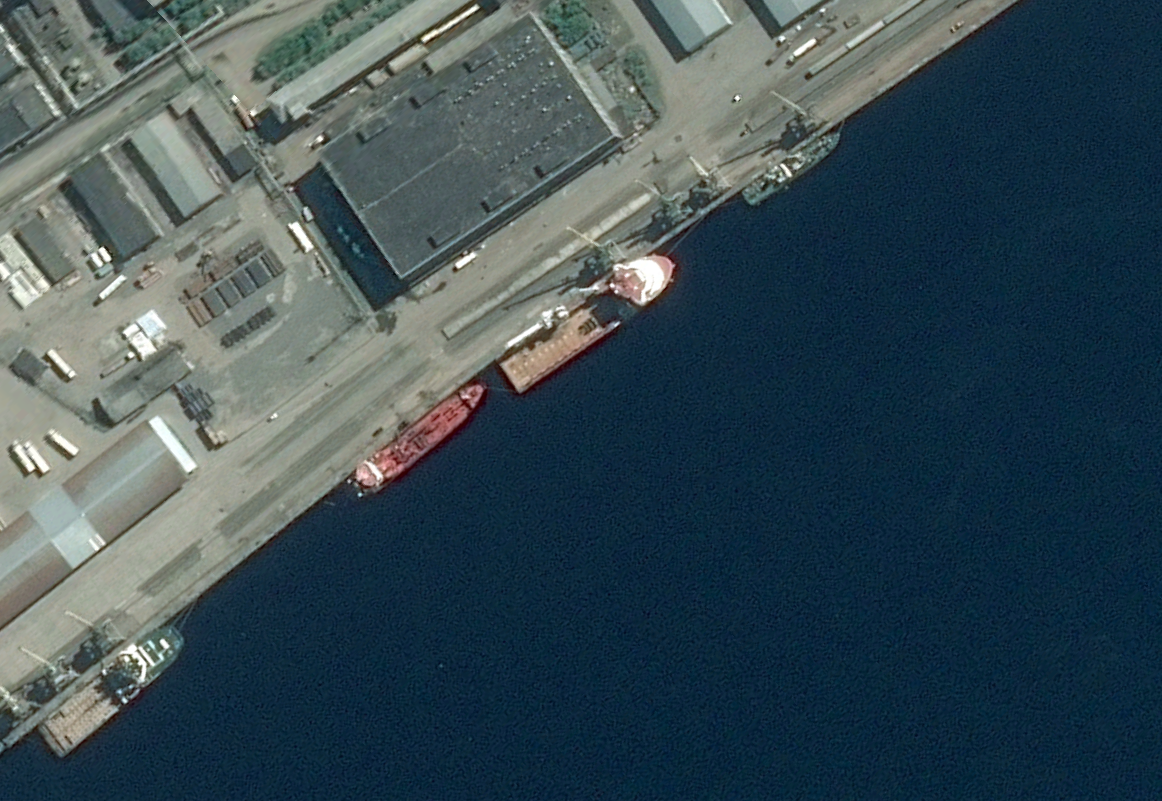}
        \includegraphics[width=\textwidth]{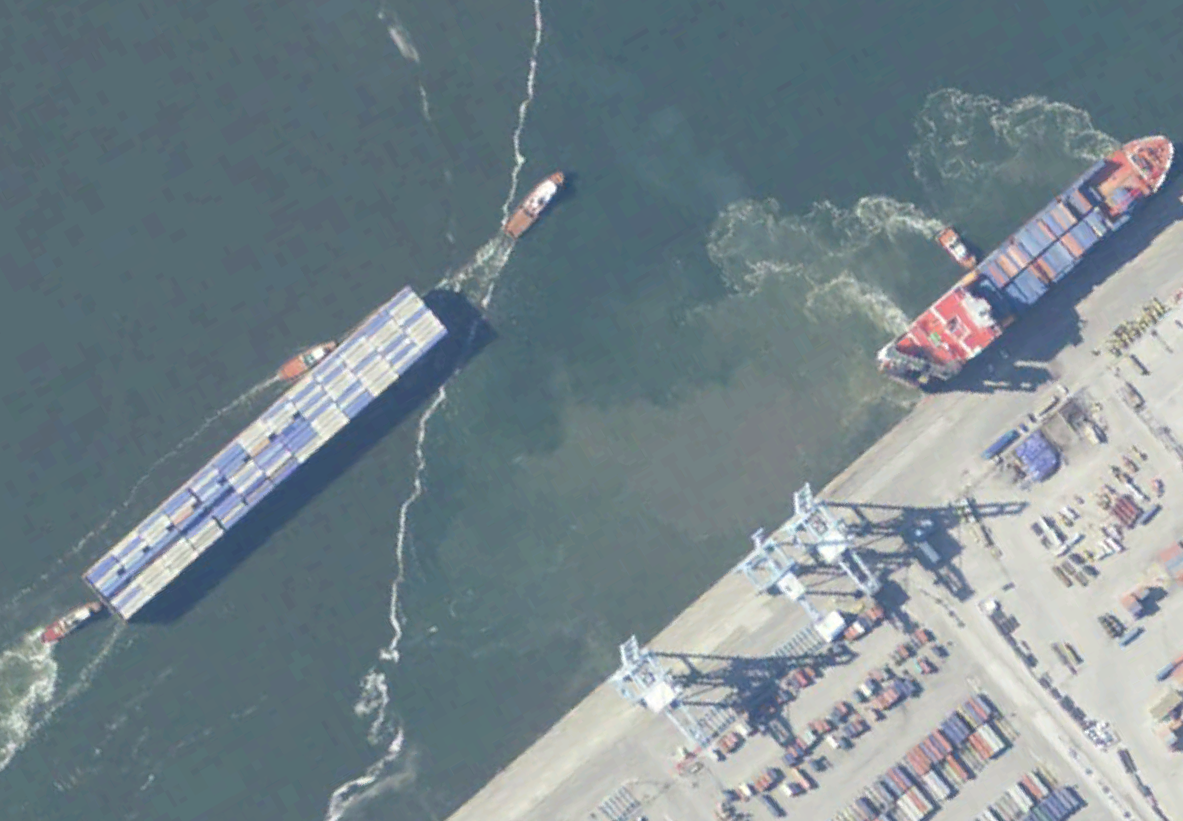}
        \includegraphics[width=\textwidth]{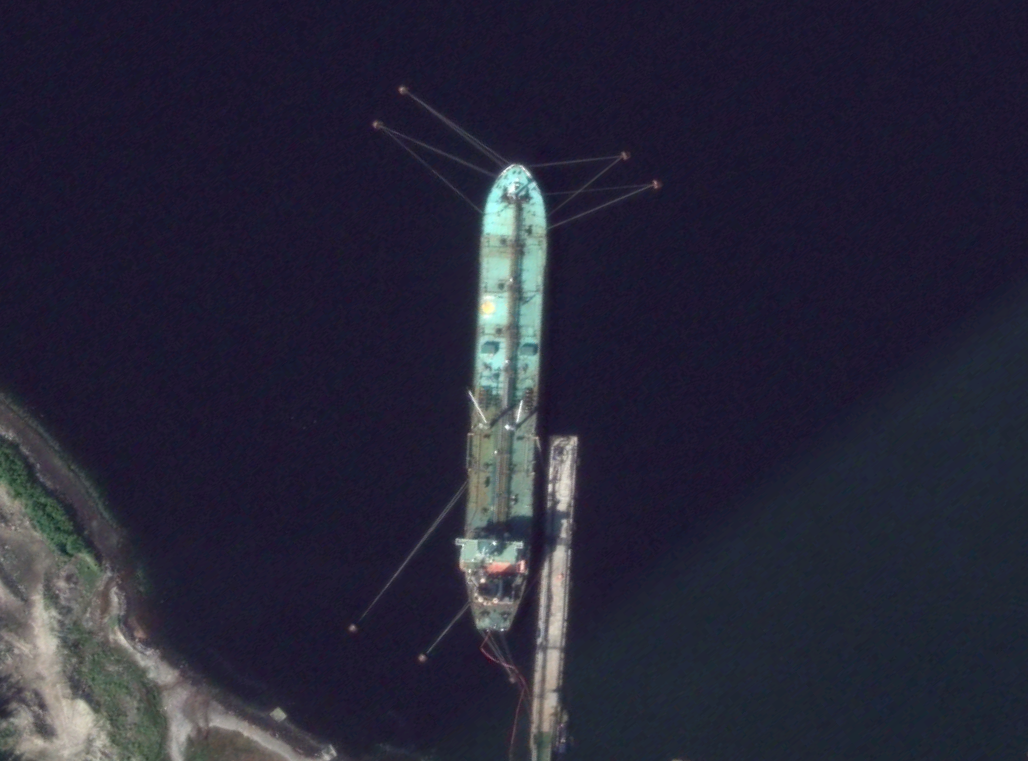}
        \vspace{3pt}
    \end{minipage}
    }%
    \subfigure[SAM annotation]{
    \begin{minipage}{0.35\linewidth} 
        \centering
        \includegraphics[width=\textwidth]{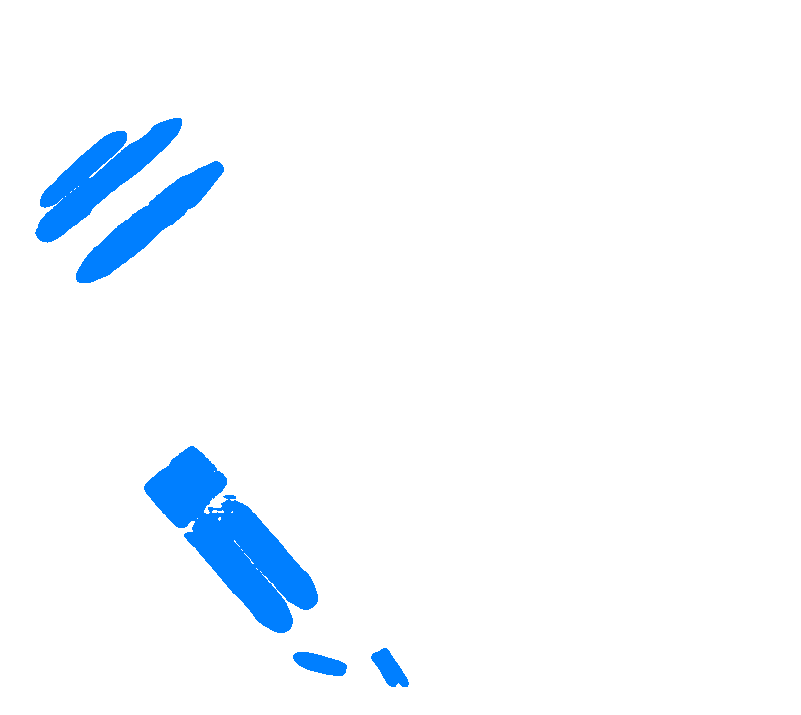}
        \includegraphics[width=\textwidth]{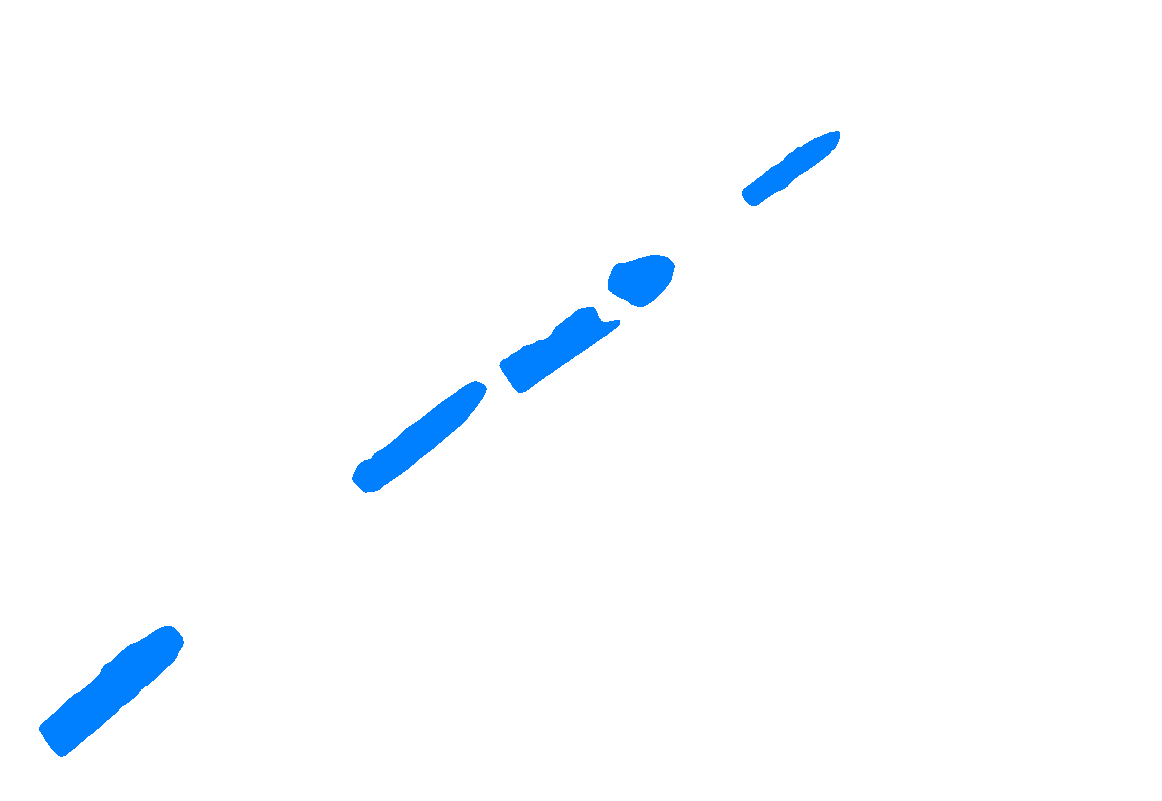}
        \includegraphics[width=\textwidth]{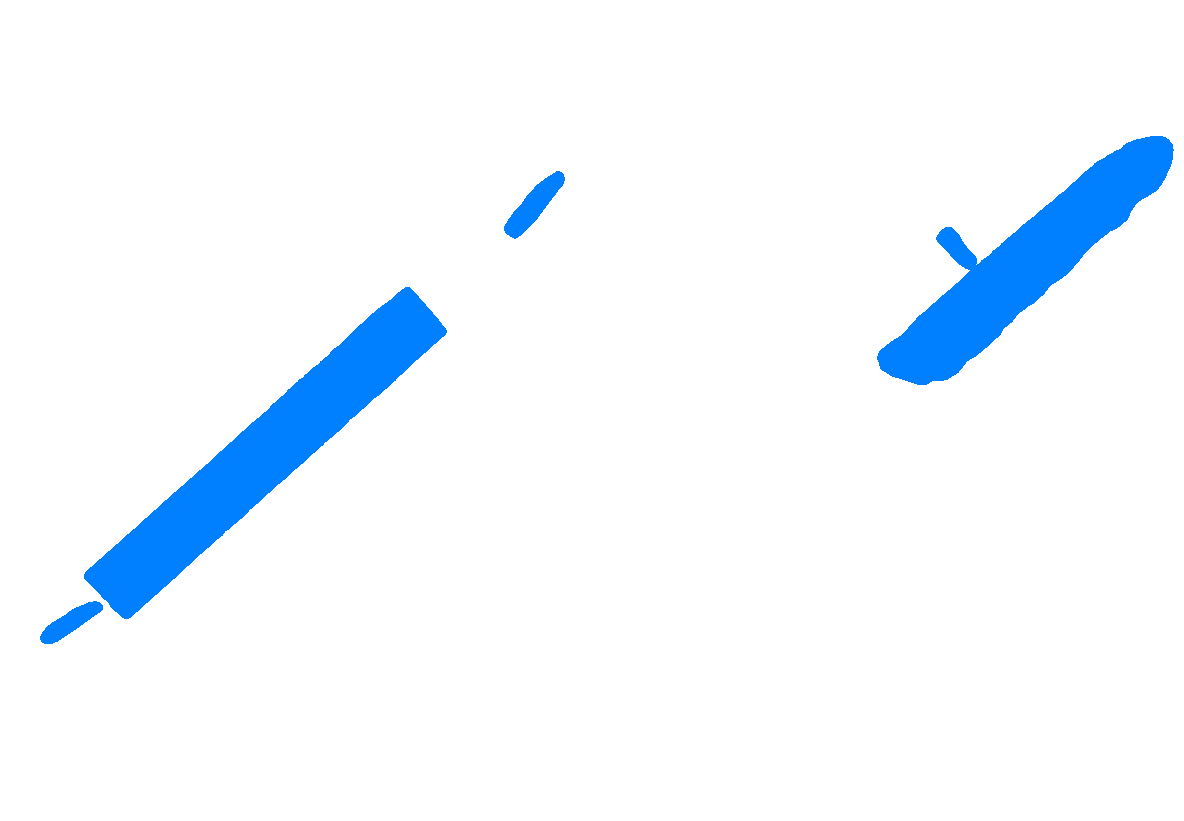}
        \includegraphics[width=\textwidth]{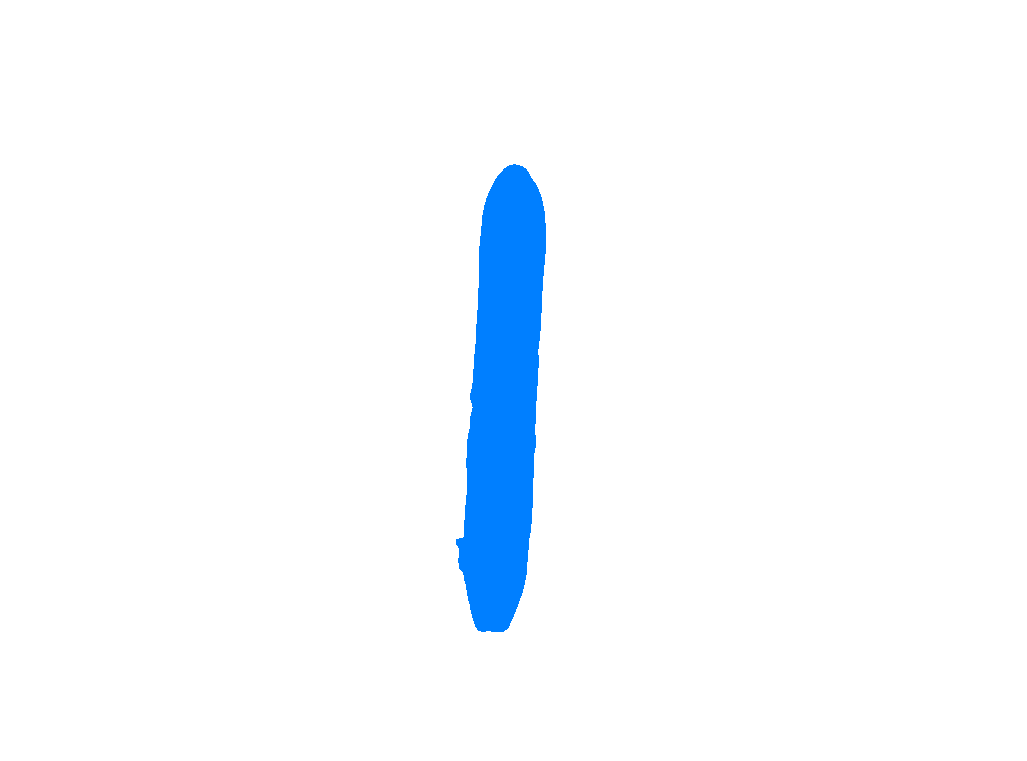}
        \vspace{3pt}
    \end{minipage}
    }%
    \centering
    \caption{Visualization of SAM-annotated samples in HRSC2016-MS}
    \label{fig_4}
\end{figure}
\begin{figure} 
    \subfigure[images]{
    \begin{minipage}[t]{0.35\linewidth} 
            \centering
            \includegraphics[width=\textwidth]{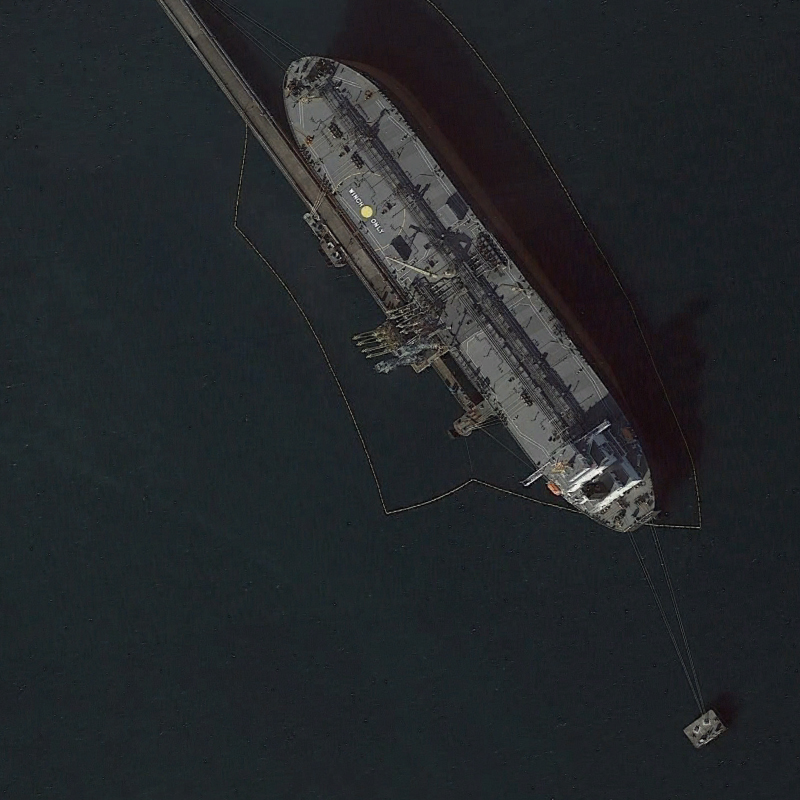}
            \includegraphics[width=\textwidth]{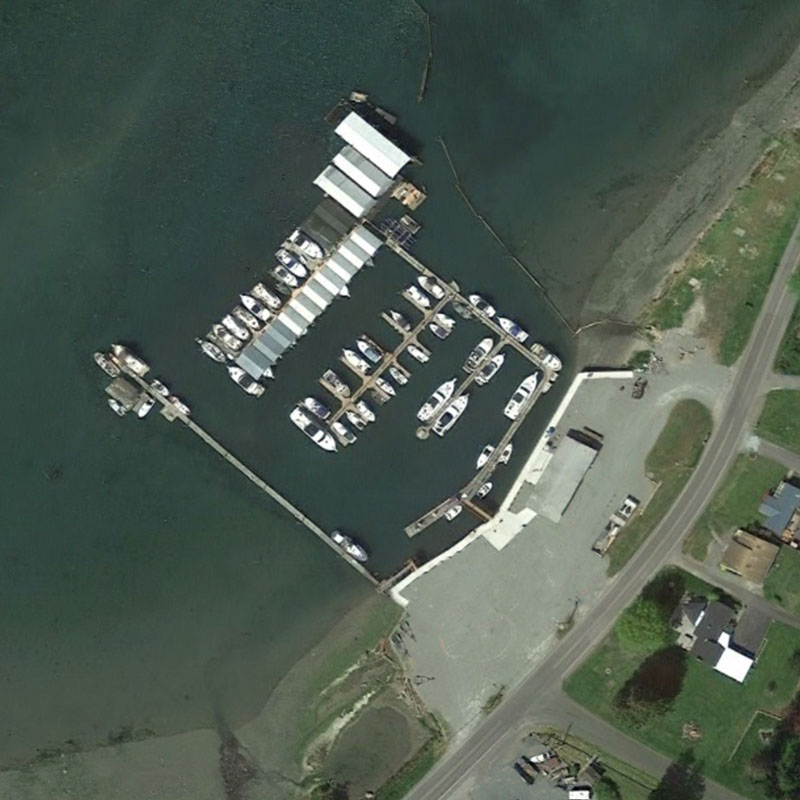}
            \includegraphics[width=\textwidth]{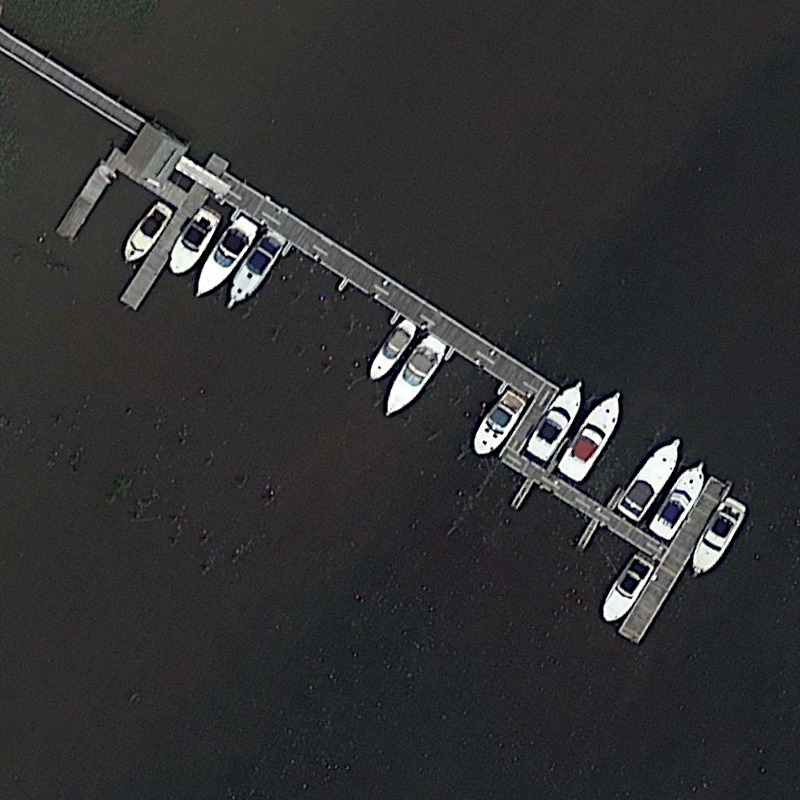}
            \vspace{3pt}
        \end{minipage}
    }%
    \subfigure[SAM annotation]{
    \begin{minipage}[t]{0.35\linewidth} 
            \centering
            \includegraphics[width=\textwidth]{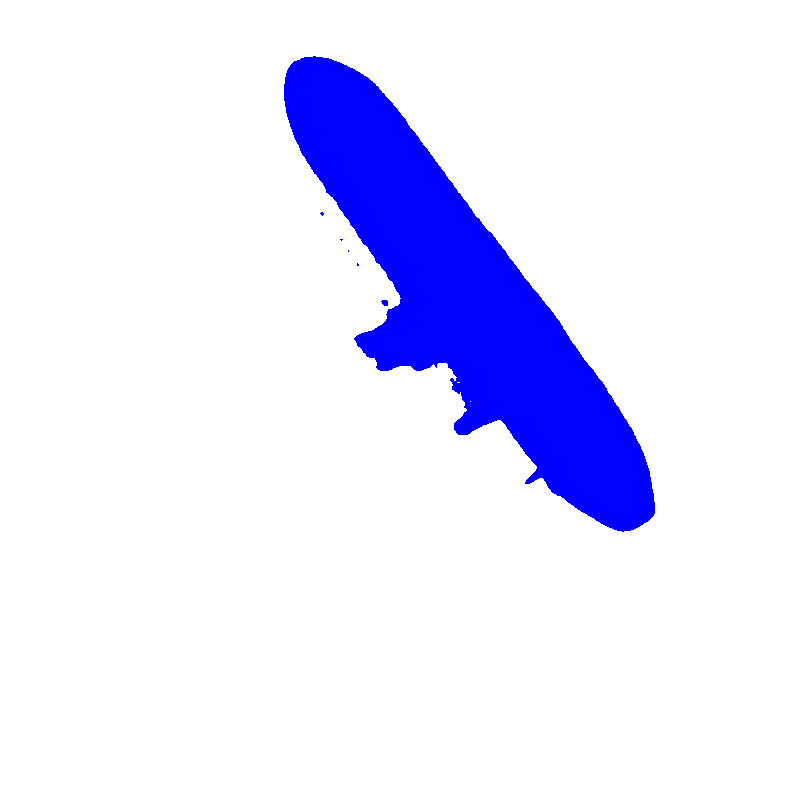}
            \includegraphics[width=\textwidth]{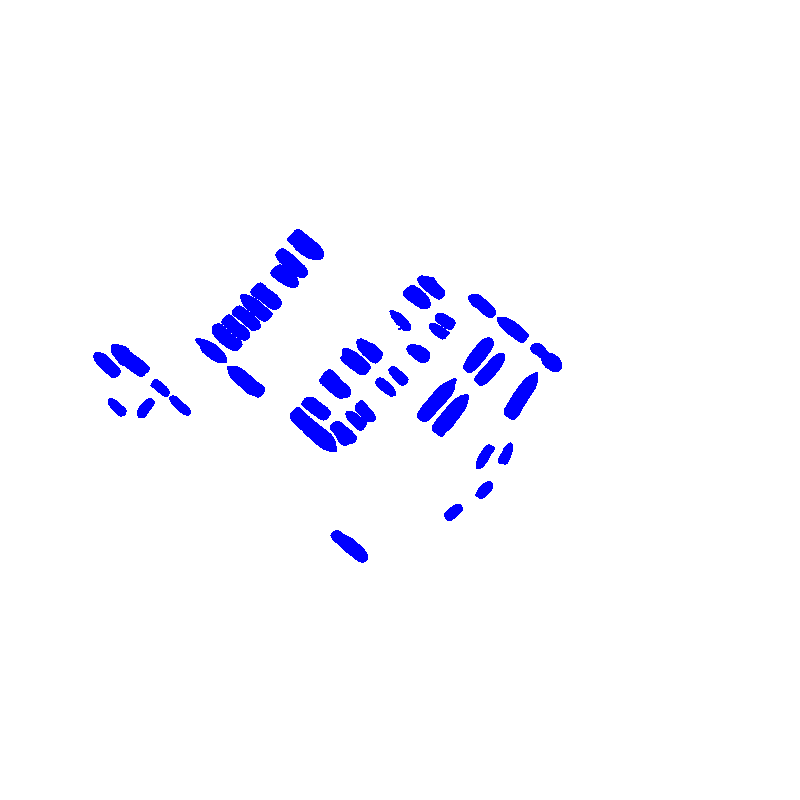}
            \includegraphics[width=\textwidth]{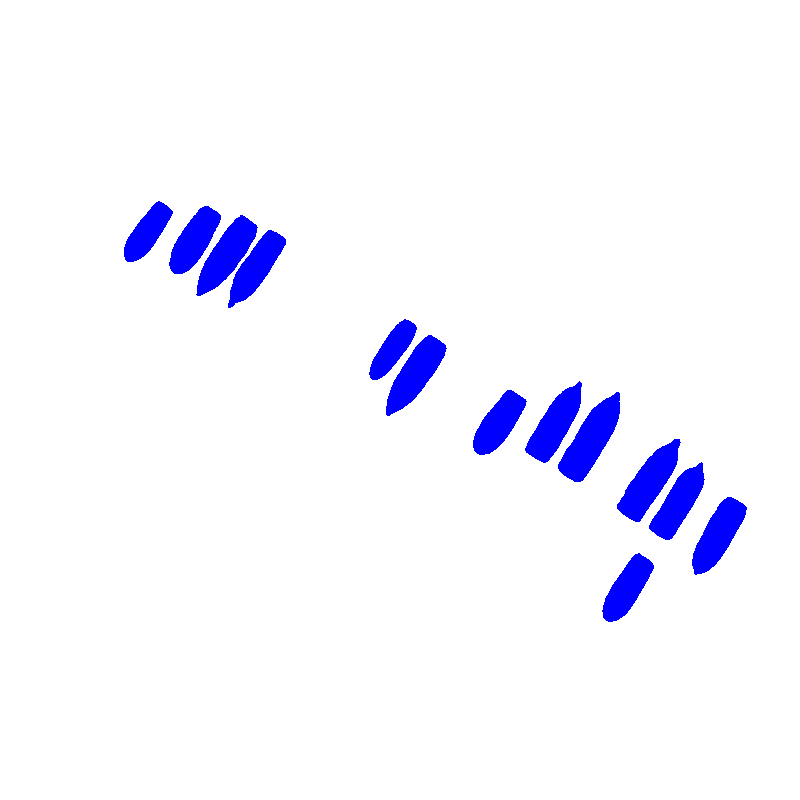}
            \vspace{3pt}
    \end{minipage}
    }%
    \centering
    \caption{Visualization of SAM-annotated samples in DIOR}
    \label{fig_5}
\end{figure}
\end{document}